\documentclass[conference]{IEEEtran}
\IEEEoverridecommandlockouts
\usepackage{cite}
\usepackage{amsmath,amssymb,amsfonts}
\usepackage{algorithmic}
\usepackage{graphicx}
\usepackage{textcomp}
\usepackage{xcolor}
\usepackage{hyperref}
\usepackage{makecell}
\usepackage{algorithm}
\usepackage{algorithmic}
\usepackage{subcaption}

\hypersetup{
    colorlinks=true, 
    linkcolor=black,  
    citecolor=black, 
    filecolor=black, 
    urlcolor=black    
}

\def\BibTeX{{\rm B\kern-.05em{\sc i\kern-.025em b}\kern-.08em
    T\kern-.1667em\lower.7ex\hbox{E}\kern-.125emX}}
\begin{document}

\title{Federated Quantum Kernel Learning for Anomaly Detection in Multivariate IoT Time-Series
\thanks{The views expressed in this article are those of the authors and do not represent the views of Wells Fargo. This article is for informational purposes only. Nothing contained in this article should be construed as investment advice. Wells Fargo makes no express or implied warranties and expressly disclaims all legal, tax, and accounting implications related to this article. }
}

\author{
\IEEEauthorblockN{
  Kuan-Cheng Chen\IEEEauthorrefmark{1}\IEEEauthorrefmark{2},
  Samuel Yen-Chi Chen\IEEEauthorrefmark{3},
  Chen-Yu Liu\IEEEauthorrefmark{4},
  Kin K. Leung\IEEEauthorrefmark{1}
}
\IEEEauthorblockA{\IEEEauthorrefmark{1}Department of Electrical and Electronic Engineering, Imperial College London, London, UK}
\IEEEauthorblockA{\IEEEauthorrefmark{2}Centre for Quantum Engineering, Science and Technology (QuEST), Imperial College London, London, UK}
\IEEEauthorblockA{\IEEEauthorrefmark{3}Wells Fargo, New York, NY, USA}
\IEEEauthorblockA{\IEEEauthorrefmark{4}Graduate Institute of Applied Physics, National Taiwan University, Taipei, Taiwan}
\thanks{Corresponding Author: kuan-cheng.chen17@imperial.ac.uk}
}


\maketitle

\begin{abstract}
The rapid growth of industrial Internet of Things (IIoT) systems has created new challenges for anomaly detection in high-dimensional, multivariate time-series, where privacy, scalability, and communication efficiency are critical. Classical federated learning approaches mitigate privacy concerns by enabling decentralized training, but they often struggle with highly non-linear decision boundaries and imbalanced anomaly distributions. To address this gap, we propose a \textbf{Federated Quantum Kernel Learning (FQKL)} framework that integrates quantum feature maps with federated aggregation to enable distributed, privacy-preserving anomaly detection across heterogeneous IoT networks. In our design, quantum edge nodes locally compute compressed kernel statistics using parameterized quantum circuits and share only these summaries with a central server, which constructs a global Gram matrix and trains a decision function (e.g., Fed-QSVM). Experimental results on synthetic IIoT benchmarks demonstrate that FQKL achieves superior generalization in capturing complex temporal correlations compared to classical federated baselines, while significantly reducing communication overhead. This work highlights the promise of quantum kernels in federated settings, advancing the path toward scalable, robust, and quantum-enhanced intelligence for next-generation IoT infrastructures.
\end{abstract}


\section{Introduction}
\label{sec:intro}
The rapid proliferation of industrial Internet of Things (IIoT) devices \cite{sisinni2018industrial} has enabled unprecedented levels of automation \cite{breivold2015internet}, monitoring \cite{civerchia2017industrial}, and optimization\cite{tran2019federated} in critical infrastructures. However, the massive volume of time-series data generated by distributed sensors, coupled with the strict requirements for data privacy and low-latency communication, presents significant challenges for centralized machine learning approaches\cite{nguyen2021federated,wang2019adaptive,zeng2021energy}. Federated learning (FL) has emerged as a promising paradigm to address these constraints by enabling collaborative training across decentralized devices without the need to share raw data \cite{konevcny2016federated}. Yet, classical federated models often struggle when confronted with complex, high-dimensional, and highly non-linear decision boundaries, which are characteristic of industrial anomaly detection scenarios\cite{zhang2024anomaly,wang2023ensuring}.

Quantum machine learning (QML) has recently attracted attention as a potential enabler of enhanced learning performance on structured and high-dimensional data \cite{dunjko2018machine, liu2025quantumrkd}. By leveraging Hilbert space embeddings and quantum feature maps, quantum kernels and variational circuits are capable of representing data correlations that are intractable for certain classical methods \cite{dunjko2018machine}. Kernel-based quantum models, such as quantum support vector machines (QSVMs)\cite{rebentrost2014quantum,chen2024quantum,chen2025validating,hsu2025quantumklstm,hsu2025quantum,chen2025compressedmediq}, have shown promise in handling non-linear classification tasks by implicitly projecting data into exponentially large feature spaces. When integrated into federated settings, such models offer a promising avenue for advancing distributed intelligence under communication, privacy, and robustness constraints\cite{ballester2025quantum,ma2025robust,chen2025consensus,chen2025noise,chen2025quantum}.

Federated quantum machine learning (FQML) \cite{chen2024introduction,chen2021federated,chehimi2022quantum} extends this paradigm by combining quantum-enhanced feature representations with classical federated aggregation strategies. In this hybrid framework \cite{mari2020transfer,chen2024quantum}, clients locally compute quantum kernel matrices or train quantum-enhanced models\cite{chen2021federated,chehimi2022quantum,liu2024federated,liu2025federated,qu2024quantum,yun2022slimmable}, which are subsequently aggregated on a central server without exposing sensitive sensor data \cite{yun2022slimmable}. This design aligns naturally with IIoT applications \cite{abou2022federated,ferrag2022edge}, where data is inherently distributed, privacy-preserving protocols are essential, and communication bandwidth is limited\cite{zafari2023resource}. Importantly, FQML does not necessitate full-scale quantum networking; rather, clients can operate quantum processors locally or via cloud backends, while the federated protocol ensures scalability and interoperability with classical infrastructure \cite{chen2021federated,chehimi2022quantum,liu2024federated,liu2025federated,qu2024quantum,yun2022slimmable}.

Despite its promise, the development of FQML for IIoT anomaly detection remains at an early stage. Key research questions include the design of quantum feature maps suitable for temporal data, the robustness of federated aggregation under client heterogeneity, and the evaluation of performance against well-established classical baselines. Furthermore, anomaly detection in IIoT often involves severe class imbalance, where conventional classifiers may yield high accuracy but poor recall. Exploring how quantum-enhanced kernels behave under such conditions, and whether they provide resilience to imbalance, is crucial for assessing the utility of quantum methods in industrial contexts.

In this work, we present a proof-of-concept study of federated quantum support vector machines (Fed-SVMs) for IIoT time-series anomaly detection. We propose a synthetic parity-phase scenario to emulate complex non-linear dependencies that are challenging for classical models. Our framework employs parameterized quantum feature maps implemented in PennyLane with up to ten qubits, coupled with federated aggregation across multiple clients. We benchmark the approach against centralized classical baselines, including SVM-RBF and Random Forest, to assess comparative performance in terms of recall, precision–recall AUC, and communication cost. The results highlight both the feasibility and the limitations of current FQML implementations, offering insights into future directions for distributed quantum–classical pipelines in industrial environments.

\begin{figure*}[t]
    \centering
    \includegraphics[width=0.95\textwidth]{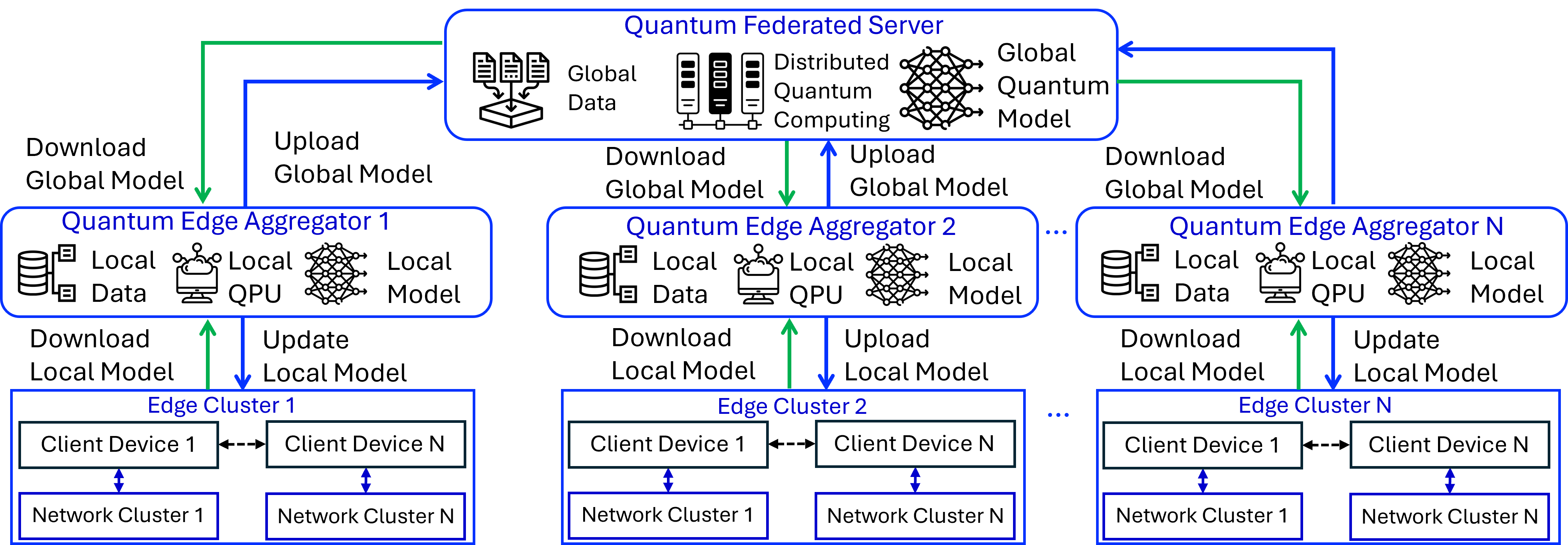}
    \caption{Proposed architecture of \textbf{Federated Quantum Kernel Learning} for anomaly detection in multivariate IoT time-series. Quantum edge nodes compute local kernel statistics using a fixed quantum feature map and share only compressed results with the \textit{Quantum Federated Server}, which aggregates them into a global Gram matrix to train a decision function (e.g., Fed-QSVM). The optimized global model is then disseminated back to all edge nodes, enabling privacy-preserving and scalable anomaly detection across heterogeneous IoT networks.}
    \label{fig:fqkl_framework}
\end{figure*}

\section{Related Work}
\label{sec:related}
\subsection{Anomaly Detection in Univariate Time Series}
Anomaly detection in univariate time series has been widely studied across domains such as finance, manufacturing, and network monitoring. Traditional statistical approaches include control chart methods (e.g., Shewhart and CUSUM tests) and autoregressive integrated moving average (ARIMA) forecasting, which identify anomalies by modeling temporal dependencies and flagging deviations beyond confidence intervals. While effective for stationary signals with well-defined distributions, these approaches often struggle with non-stationary or irregular patterns common in industrial IoT data.

Machine learning-based techniques have emerged to address such limitations. Classical supervised methods, such as one-class support vector machines (OC-SVM) \cite{hejazi2013one} and isolation forests, operate by learning normality from training data and labeling deviations as anomalies \cite{xu2023deep}. More recently, deep learning methods--particularly recurrent neural networks (RNNs) and long short-term memory networks (LSTMs) \cite{lindemann2021survey,ergen2019unsupervised}--have been applied for sequence prediction and reconstruction-based anomaly detection. Variants such as LSTM autoencoders \cite{malhotra2016lstm,said2020network} and temporal convolutional networks (TCNs) \cite{mezina2021network,he2019temporal,thill2021temporal} can capture longer temporal dependencies and adapt to complex dynamics. Despite their improved performance, these models typically require large volumes of training data and are prone to overfitting when anomalies are scarce, a condition frequently encountered in real-world IIoT scenarios.

\subsection{Anomaly Detection in Multivariate Time Series}
Multivariate time-series anomaly detection introduces additional challenges due to interdependencies across sensor modalities. Classical multivariate methods extend univariate techniques by incorporating correlation structures, for instance, through vector autoregressive (VAR) models or multivariate Gaussian processes \cite{blazquez2021review}. However, these approaches face scalability issues as the dimensionality of sensor networks grows, and often assume linear dependencies that may not hold in industrial settings.

Machine learning approaches have sought to address these limitations by leveraging representation learning. Principal component analysis (PCA) and its variants (e.g., robust PCA) are widely used for dimensionality reduction, with anomalies detected as deviations in low-rank reconstructions \cite{botterman2022robust}. Deep learning methods further extend this paradigm: LSTM and GRU-based models are employed to capture cross-sensor temporal correlations \cite{liu2020deep}, while graph neural networks (GNNs) \cite{deng2021graph} and attention-based architectures have been introduced to model sensor interdependencies more explicitly \cite{xia2023coupled}. These models have demonstrated strong performance in complex IIoT systems, yet they remain computationally demanding and may require centralized data aggregation, raising concerns around privacy and communication overhead.

\subsection{Federated Learning for Time-Series Anomaly Detection}

Federated learning \cite{konevcny2016federated} has emerged as a promising paradigm to address the privacy and communication challenges inherent in distributed industrial IoT systems. Instead of aggregating raw data from geographically distributed devices, FL enables collaborative model training through parameter sharing under secure protocols. Applied to time-series anomaly detection, FL allows heterogeneous edge devices and industrial sensors to contribute to a global model while keeping sensitive data localized.

Several works have investigated federated LSTM \cite{polap2023energy}, GRU \cite{kaur2024intrusion}, and transformer-based models \cite{luo2024securing} for anomaly detection in IIoT environments. These methods have demonstrated the ability to capture both temporal dependencies and device-specific patterns, while reducing communication costs through model compression or update sparsification. However, FL also faces challenges such as statistical heterogeneity (non-IID data across clients), system heterogeneity (varying hardware capabilities), and the potential for communication bottlenecks as models scale. Recent research has explored hybrid strategies, combining FL with model personalization \cite{pillutla2022federated,tan2022towards,li2021ditto} or meta-learning \cite{fallah2020personalized,yang2023personalized}, to mitigate performance degradation under non-uniform client distributions. Despite these advances, federated anomaly detection remains largely confined to classical ML and deep learning models, leaving open opportunities for incorporating novel paradigms such as quantum machine learning.

\section{Problem Statement}

\begin{figure}[t]
    \centering
    \includegraphics[width=0.5\textwidth]{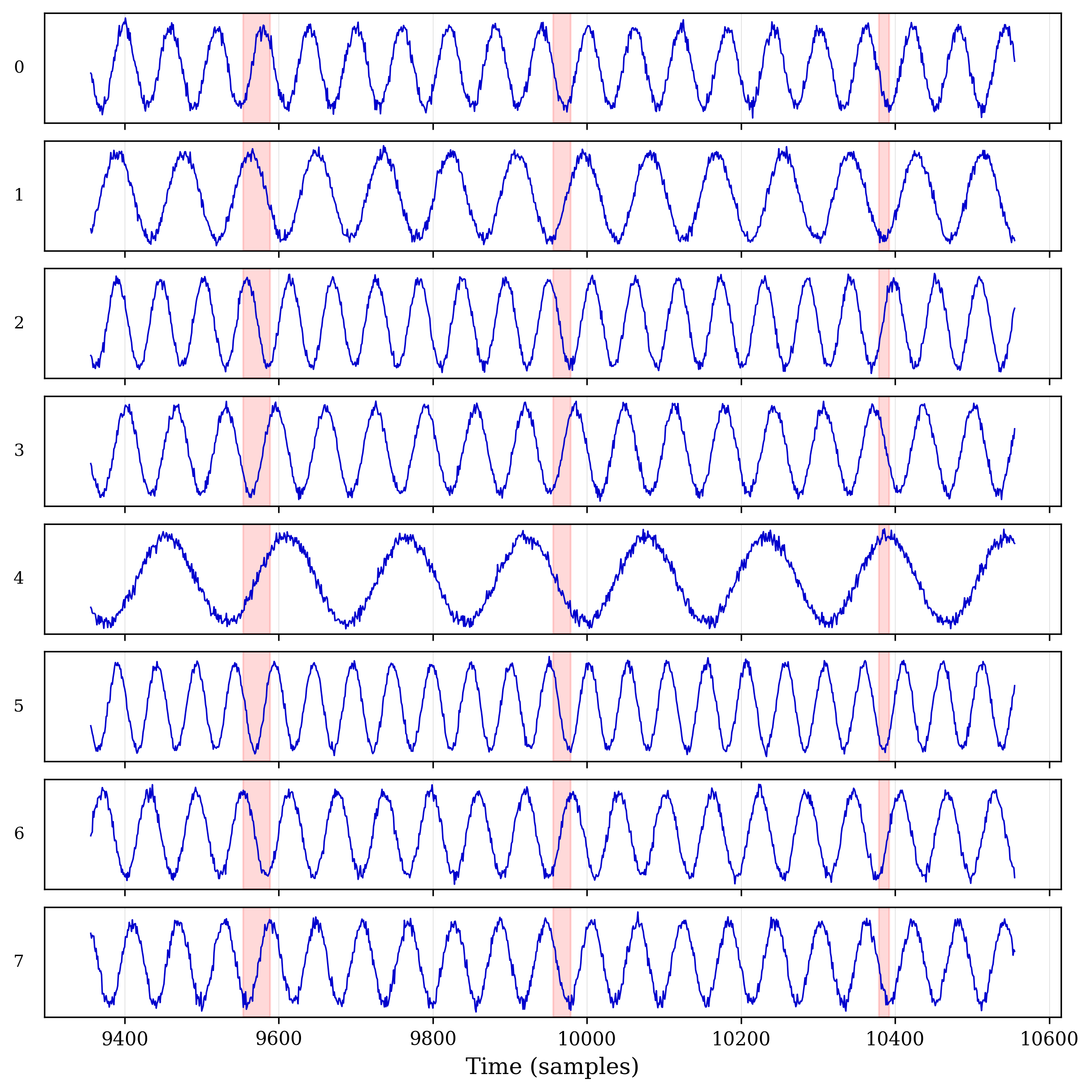}
    \caption{Multivariate IoT time-series from 8 sensors (indexed 0–7). Normal patterns are shown in blue, while anomaly intervals are highlighted in red. The aligned subplot view emphasizes temporal correlations across sensors.}
    \label{fig:data_example}
\end{figure}

\subsection{System Model}
We consider a set of $K$ IIoT clients (shown in Fig.~\ref{fig:fqkl_framework}), each equipped with heterogeneous sensors generating multivariate time-series data. The dataset on client $k \in \{1, \dots, K\}$ is denoted by
\begin{equation}
    \mathcal{D}_k = \{(x_i^{(k)}, y_i^{(k)})\}_{i=1}^{N_k},
\end{equation}
where $x_i^{(k)} \in \mathbb{R}^d$ is a $d$-dimensional feature vector obtained from a sliding window of raw sensor readings, and $y_i^{(k)} \in \{0,1\}$ is the corresponding anomaly label. The global dataset is $\mathcal{D} = \bigcup_{k=1}^K \mathcal{D}_k$, but direct sharing of $\mathcal{D}_k$ is prohibited due to privacy and bandwidth constraints.

\subsection{Dataset Setup}
To evaluate federated quantum kernel methods under realistic IIoT-inspired conditions, we adopt synthetic multivariate time-series datasets that allow explicit control of anomaly definitions while capturing the heterogeneous dynamics of industrial sensors. Each client dataset $\mathcal{D}_k$ is constructed from raw sensor streams by applying a sliding window of length $w$, yielding feature vectors $x_i^{(k)} \in \mathbb{R}^d$ with associated binary anomaly labels $y_i^{(k)} \in \{0,1\}$.

Two complementary dataset constructions are considered:

\begin{itemize}
    \item \textbf{Periodic dataset:} Sensors follow sinusoidal patterns with heterogeneous frequencies, phases, and amplitudes, perturbed by additive noise and linear mixing. Anomalies are introduced as short bursts where selected sensors exhibit abrupt drifts and transient spikes, mimicking unexpected operational faults in IIoT environments. This setup resembles practical industrial conditions such as vibration surges or sensor malfunctions.

    \item \textbf{Parity-of-Phase dataset:} Sensors again generate periodic signals with stochastic phase variations. For each sensor $j$, we compute a binary phase-change indicator
    \begin{equation}
        Z_{t,j} = \mathbb{1}\big[X_{t,j} - X_{t-\ell,j} > 0 \big],
    \end{equation}
    where $\ell$ is a lag parameter and $\mathbb{1}[\cdot]$ is the indicator function. A subset $\mathcal{S}$ of $p$ sensors is selected, and the \emph{parity signal} is defined as
    \begin{equation}
        P_t = \bigoplus_{j \in \mathcal{S}} Z_{t,j},
    \end{equation}
    where $\oplus$ denotes the XOR operation. Anomalies are triggered when $P_t=1$, and to avoid dense labeling, contiguous bursts of length $10$–$40$ samples are marked as anomalous. This construction yields a high-order, non-linear decision boundary that is challenging for classical kernel methods but naturally aligns with the representational power of entangling quantum feature maps.
\end{itemize}

\begin{figure}[t]
    \centering
    \includegraphics[width=0.5\textwidth]{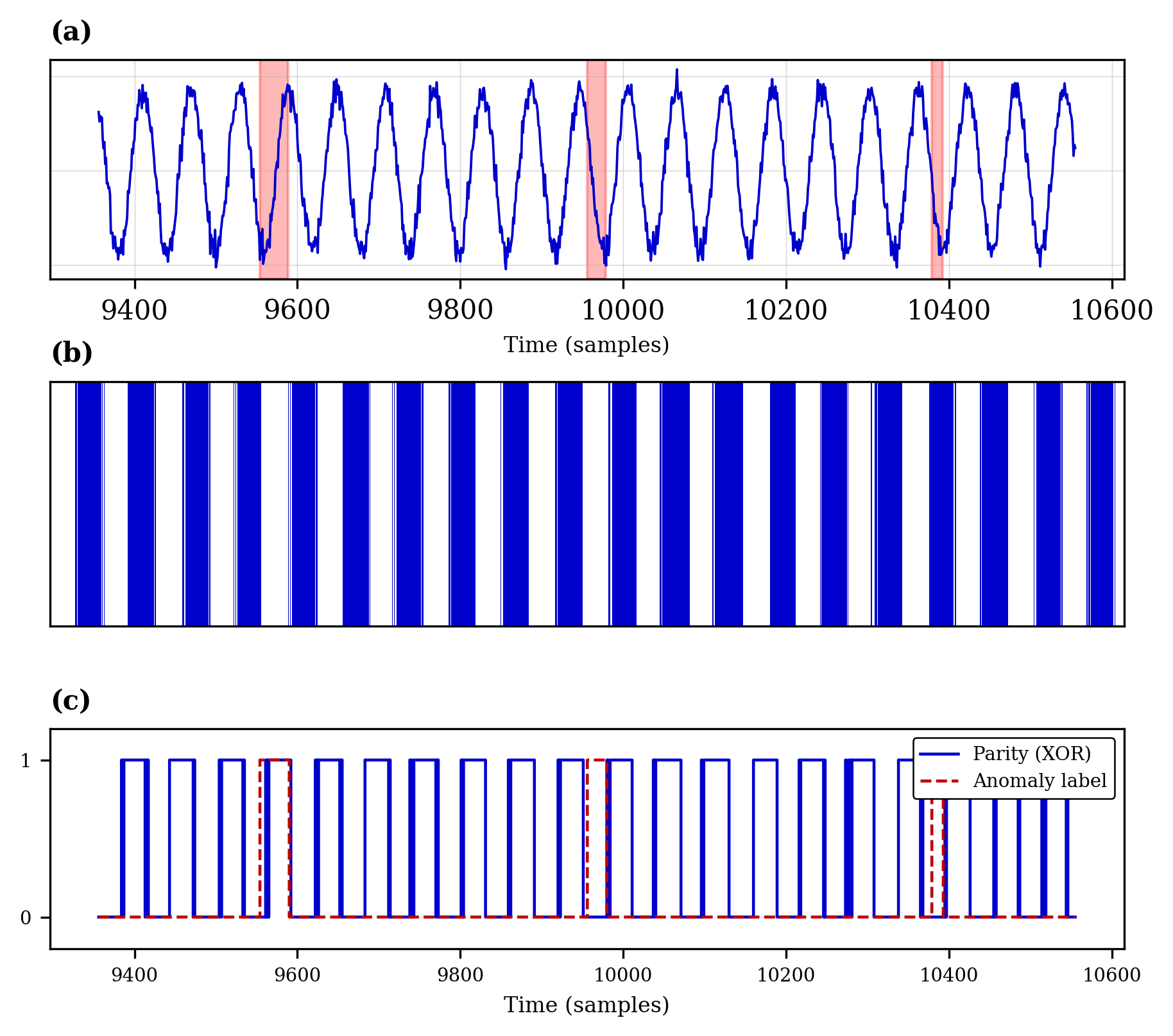}
    \caption{Illustration of the synthetic IIoT dataset. 
    (a) Example raw sensor waveforms with shaded anomaly intervals. 
    (b) Phase-change indicator matrix, where binary values capture the up/down state of selected sensors over time. 
    (c) Comparison between the derived parity (XOR of selected phase-change indicators) and the ground-truth anomaly labels. 
    Anomalies are not visible as simple amplitude outliers but instead emerge from distributed high-order correlations, motivating the use of quantum-enhanced kernels in federated IIoT anomaly detection.}
    \label{fig:iot_timeseries}
\end{figure}

\section{Preliminaries}

\subsection{Quantum Kernel Embedding}
Let $\phi: \mathbb{R}^d \to \mathcal{H}$ denote a parameterized quantum feature map that encodes a classical input $x \in \mathbb{R}^d$ into an $n$-qubit Hilbert space $\mathcal{H}$. The quantum kernel induced by $\phi$ is defined as
\begin{equation}
    \kappa(x, x') = |\langle \phi(x) | \phi(x') \rangle|^2,
\end{equation}
which measures the similarity between two time-series windows $x$ and $x'$ via their overlap in Hilbert space. The kernel matrix on client $k$ is then
\begin{equation}
    K^{(k)}_{ij} = \kappa(x_i^{(k)}, x_j^{(k)}), \quad 1 \leq i,j \leq N_k.
\end{equation}

\subsection{Local Learning Objective}
Each client $k$ trains a local support vector machine (SVM) on $(K^{(k)}, y^{(k)})$ by solving the dual optimization problem
\begin{align}
    \max_{\alpha^{(k)}} & \quad 
    \sum_{i=1}^{N_k} \alpha^{(k)}_i - \tfrac{1}{2} \sum_{i,j=1}^{N_k} \alpha^{(k)}_i \alpha^{(k)}_j y_i^{(k)} y_j^{(k)} K^{(k)}_{ij}, \\
    \text{s.t.} & \quad 0 \leq \alpha^{(k)}_i \leq C, \quad 
    \sum_{i=1}^{N_k} \alpha^{(k)}_i y_i^{(k)} = 0,
\end{align}
where $C > 0$ is the regularization parameter. The resulting model is characterized by a set of support vectors $\mathcal{S}_k = \{x_j^{(k)} : \alpha^{(k)}_j > 0\}$, dual coefficients $\alpha^{(k)}$, and bias term $b^{(k)}$.

\subsection{Federated Aggregation}
Since raw data cannot be exchanged, each client transmits only its support vectors, dual coefficients, and bias contribution to the central server. The global decision function is constructed as
\begin{equation}
    f(x) = \text{sign}\left( \sum_{k=1}^K \sum_{j \in \mathcal{S}_k} \alpha^{(k)}_j y_j^{(k)} \kappa(x_j^{(k)}, x) + b \right),
\end{equation}
where $b$ is obtained by aggregating client-specific biases, typically via weighted averaging proportional to client data sizes $N_k$.

\subsection{Problem Definition}
\textbf{Definition 1 (Federated Quantum Kernel Learning).}  
Given a set of $K$ clients with local datasets $\{\mathcal{D}_k\}_{k=1}^K$, the problem of federated quantum kernel learning is to construct a global classifier $f: \mathbb{R}^d \to \{0,1\}$ that:
\begin{enumerate}
    \item preserves data privacy by restricting communication to quantum kernel model parameters (support vectors, dual coefficients, bias);
    \item minimizes the empirical risk on the global dataset $\mathcal{D}$ under class imbalance, i.e.,
    \begin{equation}
        \min_{f \in \mathcal{F}_Q} \frac{1}{|\mathcal{D}|} \sum_{(x,y) \in \mathcal{D}} \ell(f(x), y),
    \end{equation}
    where $\ell$ is the binary classification loss and $\mathcal{F}_Q$ is the hypothesis class induced by quantum kernels; and
    \item ensures communication efficiency by bounding the payload size per client as a function of $|\mathcal{S}_k|$ rather than $N_k$.
\end{enumerate}

The core challenge is to design federated quantum kernel methods that can (i) capture highly non-linear and temporally entangled structures in IIoT time-series, (ii) remain robust to heterogeneous and imbalanced data distributions across clients, and (iii) operate within the strict bandwidth constraints typical of IIoT networks.


\subsection{Quantum States and Kernel Functions}

\subsubsection{Hilbert Space and Quantum Feature Embeddings}
A system of \(N\) qubits is described by the Hilbert space \(\mathcal{H} \cong \mathbb{C}^{2^N}\). Given a classical input \(\mathbf{x} \in \mathbb{R}^d\), a parameterized unitary embedding
\begin{equation}
U_E(\mathbf{x}): |0\rangle^{\otimes N} \mapsto |\phi(\mathbf{x})\rangle
\end{equation}
defines a quantum feature state. The corresponding density operator is
\begin{equation}
\rho(\mathbf{x}) = |\phi(\mathbf{x})\rangle \langle \phi(\mathbf{x})|,
\end{equation}
which compactly encodes the data in a potentially exponentially large Hilbert space.  

\subsubsection{Quantum Kernel Function}
The similarity between two inputs \(\mathbf{x}, \mathbf{x}'\) is quantified by the quantum kernel
\begin{equation}
K(\mathbf{x}, \mathbf{x}') = \big|\langle \phi(\mathbf{x}) | \phi(\mathbf{x}') \rangle \big|^2
= \mathrm{Tr}\!\left[\rho(\mathbf{x})\,\rho(\mathbf{x}')\right].
\end{equation}
For a dataset \(\{\mathbf{x}_i\}_{i=1}^n\), evaluating pairwise kernels yields the Gram matrix \(K \in \mathbb{R}^{n \times n}\), which is positive semidefinite and suitable for use in kernel methods such as support vector machines (SVMs) or kernel ridge regression.  

\subsection{Learning with Quantum Kernels}
In kernel SVM, the decision function for a new sample \(\mathbf{x}\) is expressed as
\begin{equation}
h(\mathbf{x}) = \mathrm{sign}\!\left( \sum_{i=1}^n \alpha_i\,y_i\,K(\mathbf{x}, \mathbf{x}_i) + b \right),
\end{equation}
where \(\alpha_i\) and \(b\) are determined by solving the dual optimization problem
\begin{equation}
\max_{\boldsymbol{\alpha}} \; \mathbf{1}^\top \boldsymbol{\alpha} 
- \tfrac{1}{2} \boldsymbol{\alpha}^\top \big( Y K Y \big) \boldsymbol{\alpha}, 
\quad \text{s.t. } \; 0 \leq \alpha_i \leq C,\; \sum_i \alpha_i y_i = 0,
\end{equation}
with \(Y = \mathrm{diag}(y_1, \ldots, y_n)\). Here, the kernel matrix \(K\) may be generated by a quantum device, while the optimization remains classical.

\subsection{Federated Quantum Kernel Learning}

In federated learning, training data are distributed across \(M\) clients \(\{\mathcal{D}_j\}_{j=1}^M\), each holding local samples \(\mathcal{D}_j = \{(\mathbf{x}_i^{(j)}, y_i^{(j)})\}\). Directly pooling raw data may be infeasible due to privacy or communication constraints. Instead, each client computes its local quantum kernel evaluations:
\begin{equation}
K_j(\mathbf{x}, \mathbf{x}') = 
\big|\langle \phi(\mathbf{x}) | \phi(\mathbf{x}') \rangle \big|^2, 
\quad \mathbf{x},\mathbf{x}' \in \mathcal{D}_j.
\end{equation}

The clients then share compressed kernel statistics (e.g., support vectors, partial kernel submatrices, or aggregated feature maps) with a central server. The server aggregates the information into a global kernel matrix:
\begin{equation}
K_{\mathrm{global}} = \frac{1}{M} \sum_{j=1}^M \Pi_j \, K_j \, \Pi_j^\top,
\end{equation}
where \(\Pi_j\) is an embedding operator aligning the indices of client \(j\) into the global kernel space. This ensures that the global kernel preserves contributions from all clients without requiring raw data exchange.

Finally, the global kernel matrix \(K_{\mathrm{global}}\) is used in the classical optimization step of SVM (or other kernel methods), yielding a decision function:
\begin{equation}
h(\mathbf{x}) = \mathrm{sign}\!\left( \sum_{j=1}^M \sum_{\mathbf{x}_i^{(j)} \in \mathcal{D}_j} 
\alpha_i^{(j)}\,y_i^{(j)}\,K(\mathbf{x}, \mathbf{x}_i^{(j)}) + b \right).
\end{equation}
This formulation integrates quantum kernel embeddings with federated aggregation, enabling distributed and privacy-preserving quantum-enhanced learning across heterogeneous IoT clients.

\subsection{Federated Quantum Kernel Learning}

We propose a federated learning framework for quantum kernel methods, in which multiple clients collaboratively train a global model without sharing raw data. Consider $M$ clients, each holding a local dataset $\mathcal{D}_j = \{(\mathbf{x}_i^{(j)}, y_i^{(j)})\}_{i=1}^{n_j}$. Each client is capable of evaluating quantum kernel overlaps using the fixed feature map $U_E(\cdot)$ defined in Section~II. This yields local kernel matrices $K_j \in \mathbb{R}^{n_j \times n_j}$ associated with its private dataset.  

Rather than transmitting local samples, each client compresses its kernel information---either through support vector selection or by reporting submatrices aligned with the global index set---and sends only these compressed statistics to the server. This strategy ensures that sensitive local data remain private, while still contributing to the global learning task.

\subsection{Global Aggregation and Optimization}

At the server, local contributions are assembled into a global Gram matrix via index-alignment operators $\Pi_j$:  
\begin{equation}
K_{\text{global}} = \frac{1}{M} \sum_{j=1}^M \Pi_j K_j \Pi_j^\top.
\end{equation}
The server then solves the standard support vector machine dual problem over $K_{\text{global}}$:  
\begin{equation}
\max_{\boldsymbol{\alpha}} \quad \mathbf{1}^\top \boldsymbol{\alpha} - \tfrac{1}{2}\boldsymbol{\alpha}^\top (Y K_{\text{global}} Y)\boldsymbol{\alpha},
\quad \text{s.t. } 0 \leq \alpha_i \leq C,\; \sum_i \alpha_i y_i = 0,
\end{equation}
where $Y = \mathrm{diag}(y_1,\ldots,y_n)$ encodes labels from all clients. Solving this optimization yields the global dual coefficients $\{\alpha_i\}$ and bias $b$.  

The resulting global decision function
\begin{equation}
h(\mathbf{x}) = \mathrm{sign}\!\left(\sum_{i=1}^n \alpha_i y_i K(\mathbf{x}, \mathbf{x}_i) + b\right)
\end{equation}
is then broadcast to all clients. Each client can apply $h(\cdot)$ locally for inference without requiring access to other clients’ data. This architecture provides communication efficiency and privacy guarantees, making it particularly suited to federated IoT deployments where both quantum resources and communication bandwidth are constrained.

\begin{figure*}[!t]
    \centering
    \begin{subfigure}[!t]{0.32\textwidth}
        \centering
        \includegraphics[width=\linewidth]{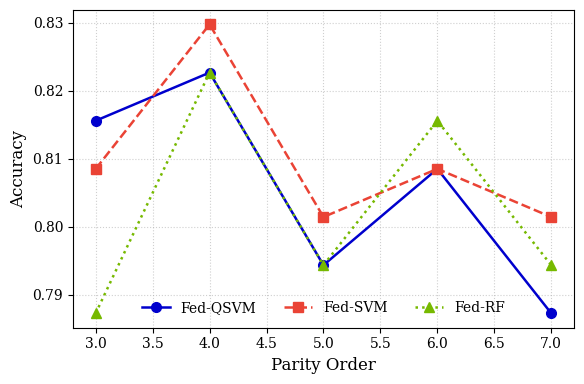}
        \caption{Impact of parity order on accuracy.}
        \label{fig:parity_order}
    \end{subfigure}
    \hfill
    \begin{subfigure}[!t]{0.32\textwidth}
        \centering
        \includegraphics[width=\linewidth]{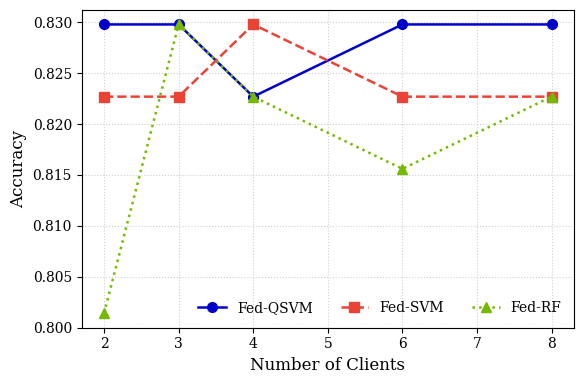}
        \caption{Scalability with number of clients.}
        \label{fig:clients}
    \end{subfigure}
    \hfill
    \begin{subfigure}[!t]{0.32\textwidth}
        \centering
        \includegraphics[width=\linewidth]{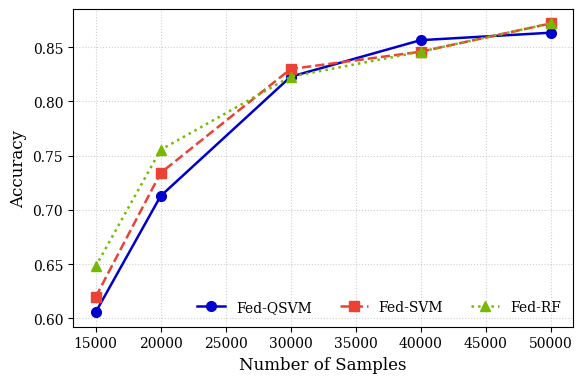}
        \caption{Effect of training data size.}
        \label{fig:num_samples}
    \end{subfigure}

    \caption{Benchmarking of federated quantum SVM (Fed-QSVM) against classical baselines (Fed-SVM, Fed-RF). (a) Fed-QSVM sustains accuracy under increasing parity order, unlike classical models that degrade with higher feature interactions. (b) Fed-QSVM remains robust with growing client numbers, whereas Fed-RF shows higher variance and Fed-SVM degrades moderately. (c) While requiring more samples in the low-data regime, Fed-QSVM surpasses classical baselines with larger datasets, demonstrating superior scalability in distributed IoT anomaly detection.}

    \label{fig:benchmark_results}
\end{figure*}

\subsection{Federated Quantum Kernel Learning Algorithm}

Algorithm~\ref{alg:fqkl} summarizes the FQKL procedure. Each client computes its local kernel information, compresses and transmits it to the server, where aggregation and global optimization are performed. The global model is then redistributed for inference. This workflow decouples expensive quantum kernel evaluation (executed in parallel across clients) from the centralized classical optimization (executed once at the server), thereby balancing resource usage across the federation.

\begin{algorithm}[h]
\caption{FQKL with Support-Vector Sparsification and Quantization}
\label{alg:fqkl}
\begin{algorithmic}[1]
\STATE \textbf{Input:} clients $\{\mathcal{D}_j\}_{j=1}^M$, feature map $U_E(\cdot)$, budget $m$ SVs per client, $b$-bit quantization
\FOR{each client $j$ in parallel}
  \STATE Compute local Gram $K_j$ by PQC overlaps with $S$ shots
  \STATE Train local SVM on $(K_j, y^{(j)})$; obtain $(\alpha^{(j)}, b^{(j)})$
  \STATE Keep top-$m$ support vectors by $|\alpha^{(j)}_i|$; quantize $\alpha^{(j)}$ to $b$ bits
  \STATE Transmit $\{(\mathrm{id}_i, \mathrm{hash}(x_i^{(j)})), \mathrm{quant}(\alpha^{(j)}_i), y^{(j)}_i\}_{i\in\mathcal{S}_j}$ and $b^{(j)}$
\ENDFOR
\STATE Server reconstructs block-sparse $K_{\text{global}}$ via on-demand overlaps
\STATE Solve global SVM dual; obtain $\{\alpha_i\}, b$
\STATE Broadcast $\{(\mathrm{id}_i, \alpha_i, y_i)\}$ and $b$ to all clients
\end{algorithmic}
\end{algorithm}


\section{Experiments}
\label{sec:experiments}
\subsection{Benchmark of Federated Learning}
The results obtained from the parity order experiments (Fig.~\ref{fig:benchmark_results}(a)) clearly highlight the trade-off between problem complexity and classification accuracy across federated models. As expected, increasing the parity order, which imposes a higher non-linear dependency among the input features, leads to a general decline in accuracy across all methods. Notably, Fed-QSVM demonstrates strong resilience when compared with its classical counterparts, maintaining stable accuracy even as parity order increases from 3 to 7. By contrast, Fed-SVM exhibits more pronounced performance fluctuations, while Fed-RF, although competitive at lower orders, suffers from instability at higher complexities. This indicates that quantum-enhanced kernels provide superior generalization in high-dimensional feature interactions, which is critical for anomaly detection in IoT sensor networks where signal correlations are inherently complex.

The analysis of client scalability (Fig.~\ref{fig:benchmark_results}(b)) reveals another important dimension of federated learning in distributed IoT environments. Fed-QSVM achieves consistently high performance regardless of the number of participating clients, suggesting robustness against data fragmentation and heterogeneous distributions. Classical baselines, in particular Fed-RF, exhibit larger variance as the number of clients increases, highlighting their sensitivity to partitioned training data. Fed-SVM remains competitive but shows slight degradation when scaling beyond moderate client counts. The stability of Fed-QSVM in this federated setting emphasizes the advantage of leveraging quantum kernel methods, which are less reliant on centralized data aggregation and more effective in extracting non-local correlations across distributed clients — a key advantage for large-scale IoT infrastructures.

Finally, varying the number of training samples (Fig.~\ref{fig:benchmark_results}(c)) demonstrates the asymptotic behavior of the models with respect to data availability. In the low-data regime (15k–20k samples), Fed-QSVM initially underperforms classical baselines, reflecting the higher sample complexity often required by quantum kernel models. However, as the dataset size increases, Fed-QSVM rapidly closes the gap and converges to comparable or superior accuracy at 40k–50k samples. This observation suggests that while quantum-enhanced methods may incur an initial overhead in data requirements, they exhibit stronger scalability and improved generalization with sufficient training samples. For IoT applications where data collection can be continuous and abundant, this property positions Fed-QSVM as a promising candidate for future quantum-classical hybrid pipelines, offering long-term advantages in accuracy and resilience over purely classical federated models.

\subsection{Communication–Accuracy Trade-off in Federated Learning}

\begin{figure}[!t]
    \centering
    \includegraphics[width=0.48\textwidth]{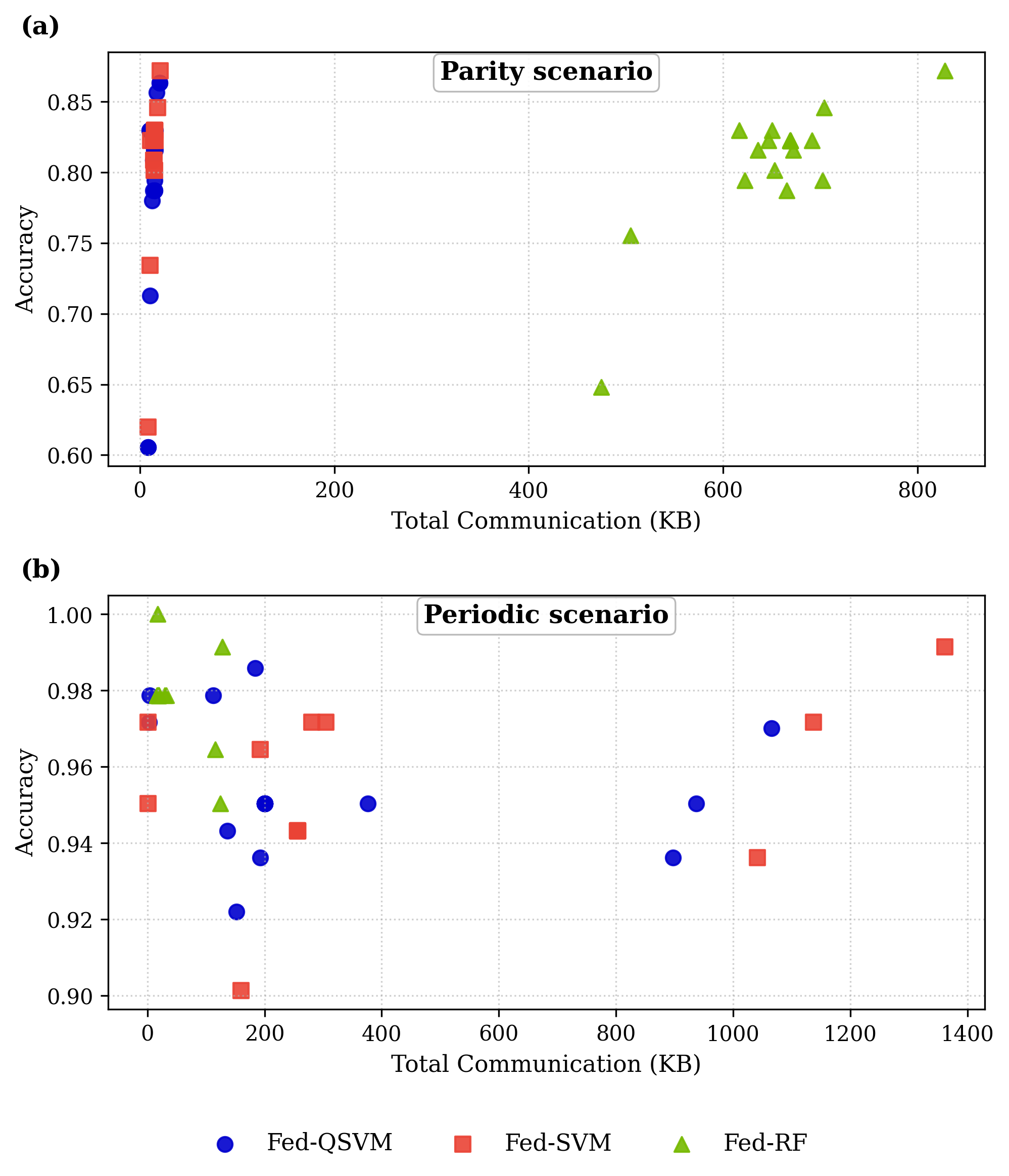}
    \caption{Communication–accuracy trade-off in federated learning. 
(a) Parity scenario: quantum kernels offer clear communication efficiency over classical baselines. 
(b) Periodic scenario: all models achieve high accuracy with overlapping communication ranges.}
    \label{fig:comm_tradeoff}
\end{figure}

The relationship between communication overhead and model accuracy is critical in federated IoT environments, where bandwidth and latency constraints are often dominant bottlenecks. Fig.~\ref{fig:comm_tradeoff} illustrates this trade-off by comparing Fed-QSVM against classical baselines (Fed-SVM and Fed-RF) under both parity and periodic scenarios.

In the parity scenario, Fed-QSVM and Fed-SVM achieve competitive accuracy while requiring only a few kilobytes of communication, in stark contrast to Fed-RF, which incurs orders of magnitude greater communication cost to reach similar or slightly higher accuracy. This clearly highlights the communication efficiency of kernel-based federated models, with Fed-QSVM demonstrating robustness in capturing high-order feature interactions without incurring excessive exchange overhead. By contrast, Fed-RF’s reliance on ensemble aggregation significantly amplifies its communication demand, making it less suited for bandwidth-limited IoT deployments.

The periodic scenario presents a more nuanced picture. All three methods achieve high accuracy (exceeding 0.9), and the communication costs, while different, show partial overlap. In this case, the advantage of Fed-QSVM is less pronounced, as the simpler periodic patterns are sufficiently captured by classical baselines. These results suggest that the primary benefit of quantum kernels emerges in tasks characterized by complex, non-linear correlations (e.g., parity anomalies), where they offer a superior balance between communication cost and learning performance.

\section{Conclusion}
\label{sec:conclusion}
In summary, this work presented the FQKL framework for anomaly detection in multivariate IoT time-series data, integrating quantum kernel methods within a federated learning architecture to enable collaborative model training without compromising data privacy. Systematic experiments across diverse data scales, client configurations, and quantum resource settings demonstrated that FQKL achieves competitive or superior performance relative to classical federated baselines such as SVM and Random Forest, benefiting from the enhanced representational capacity of quantum feature maps in capturing complex temporal correlations. The framework further proved scalable and communication-efficient, leveraging compressed kernel statistics to minimize overhead in resource-constrained IoT environments. By aligning with the emerging paradigm of distributed quantum–classical hybrid systems, FQKL establishes a foundation for future exploration in quantum-enhanced federated learning, including adaptive quantum feature design, efficient support vector selection, and implementation on near-term quantum hardware—advancing the realization of scalable, privacy-preserving, and intelligent IoT networks.

\section*{Acknowledgment}
This work was supported by the Engineering and Physical Sciences Research Council (EPSRC) under grant number EP/W032643/1.



\bibliographystyle{ieeetr}
\bibliography{references}

\begin{thebibliography}{10}

\bibitem{sisinni2018industrial}
E.~Sisinni, A.~Saifullah, S.~Han, U.~Jennehag, and M.~Gidlund, ``Industrial internet of things: Challenges, opportunities, and directions,'' {\em IEEE transactions on industrial informatics}, vol.~14, no.~11, pp.~4724--4734, 2018.

\bibitem{breivold2015internet}
H.~P. Breivold and K.~Sandstr{\"o}m, ``Internet of things for industrial automation--challenges and technical solutions,'' in {\em 2015 IEEE International Conference on Data Science and Data Intensive Systems}, pp.~532--539, IEEE, 2015.

\bibitem{civerchia2017industrial}
F.~Civerchia, S.~Bocchino, C.~Salvadori, E.~Rossi, L.~Maggiani, and M.~Petracca, ``Industrial internet of things monitoring solution for advanced predictive maintenance applications,'' {\em Journal of Industrial Information Integration}, vol.~7, pp.~4--12, 2017.

\bibitem{tran2019federated}
N.~H. Tran, W.~Bao, A.~Zomaya, M.~N. Nguyen, and C.~S. Hong, ``Federated learning over wireless networks: Optimization model design and analysis,'' in {\em IEEE INFOCOM 2019-IEEE conference on computer communications}, pp.~1387--1395, IEEE, 2019.

\bibitem{nguyen2021federated}
D.~C. Nguyen, M.~Ding, P.~N. Pathirana, A.~Seneviratne, J.~Li, and H.~V. Poor, ``Federated learning for internet of things: A comprehensive survey,'' {\em IEEE communications surveys \& tutorials}, vol.~23, no.~3, pp.~1622--1658, 2021.

\bibitem{wang2019adaptive}
S.~Wang, T.~Tuor, T.~Salonidis, K.~K. Leung, C.~Makaya, T.~He, and K.~Chan, ``Adaptive federated learning in resource constrained edge computing systems,'' {\em IEEE journal on selected areas in communications}, vol.~37, no.~6, pp.~1205--1221, 2019.

\bibitem{zeng2021energy}
Q.~Zeng, Y.~Du, K.~Huang, and K.~K. Leung, ``Energy-efficient resource management for federated edge learning with cpu-gpu heterogeneous computing,'' {\em IEEE Transactions on Wireless Communications}, vol.~20, no.~12, pp.~7947--7962, 2021.

\bibitem{konevcny2016federated}
J.~Kone{\v{c}}n{\`y}, H.~B. McMahan, F.~X. Yu, P.~Richt{\'a}rik, A.~T. Suresh, and D.~Bacon, ``Federated learning: Strategies for improving communication efficiency,'' {\em arXiv preprint arXiv:1610.05492}, 2016.

\bibitem{zhang2024anomaly}
C.~Zhang, S.~Yang, L.~Mao, and H.~Ning, ``Anomaly detection and defense techniques in federated learning: a comprehensive review,'' {\em Artificial Intelligence Review}, vol.~57, no.~6, p.~150, 2024.

\bibitem{wang2023ensuring}
H.~Wang, C.~H. Liu, H.~Yang, G.~Wang, and K.~K. Leung, ``Ensuring threshold aoi for uav-assisted mobile crowdsensing by multi-agent deep reinforcement learning with transformer,'' {\em IEEE/ACM Transactions on Networking}, vol.~32, no.~1, pp.~566--581, 2023.

\bibitem{dunjko2018machine}
V.~Dunjko and H.~J. Briegel, ``Machine learning \& artificial intelligence in the quantum domain: a review of recent progress,'' {\em Reports on Progress in Physics}, vol.~81, no.~7, p.~074001, 2018.

\bibitem{liu2025quantumrkd}
C.-Y. Liu, K.-C. Chen, K.~Murota, S.~Y.-C. Chen, and E.~Rinaldi, ``Quantum relational knowledge distillation,'' {\em arXiv preprint arXiv:2508.13054}, 2025.

\bibitem{rebentrost2014quantum}
P.~Rebentrost, M.~Mohseni, and S.~Lloyd, ``Quantum support vector machine for big data classification,'' {\em Physical review letters}, vol.~113, no.~13, p.~130503, 2014.

\bibitem{chen2024quantum}
K.-C. Chen, X.~Xu, H.~Makhanov, H.-H. Chung, and C.-Y. Liu, ``Quantum-enhanced support vector machine for large-scale multi-class stellar classification,'' in {\em International Conference on Intelligent Computing}, pp.~155--168, Springer, 2024.

\bibitem{chen2025validating}
K.-C. Chen, T.-Y. Li, Y.-Y. Wang, S.~See, C.-C. Wang, R.~Wille, N.-Y. Chen, A.-C. Yang, and C.-Y. Lin, ``Validating large-scale quantum machine learning: Efficient simulation of quantum support vector machines using tensor networks,'' {\em Machine Learning: Science and Technology}, vol.~6, no.~1, p.~015047, 2025.

\bibitem{hsu2025quantumklstm}
Y.-C. Hsu, T.-Y. Li, and K.-C. Chen, ``Quantum kernel-based long short-term memory,'' in {\em 2025 IEEE International Conference on Acoustics, Speech, and Signal Processing Workshops (ICASSPW)}, pp.~1--5, IEEE, 2025.

\bibitem{hsu2025quantum}
Y.-C. Hsu, N.-Y. Chen, T.-Y. Li, P.-H.~H. Lee, and K.-C. Chen, ``Quantum kernel-based long short-term memory for climate time-series forecasting,'' in {\em 2025 International Conference on Quantum Communications, Networking, and Computing (QCNC)}, pp.~421--426, IEEE, 2025.

\bibitem{chen2025compressedmediq}
K.-C. Chen, Y.-T. Li, T.-Y. Li, C.-Y. Liu, P.-H.~H. Lee, and C.-Y. Chen, ``Compressedmediq: Hybrid quantum machine learning pipeline for high-dimensional neuroimaging data,'' in {\em 2025 IEEE International Conference on Acoustics, Speech, and Signal Processing Workshops (ICASSPW)}, pp.~1--5, IEEE, 2025.

\bibitem{ballester2025quantum}
R.~Ballester, J.~Cerquides, and L.~Artiles, ``Quantum federated learning: a comprehensive literature review of foundations, challenges, and future directions,'' {\em Quantum Machine Intelligence}, vol.~7, no.~2, pp.~1--29, 2025.

\bibitem{ma2025robust}
W.~Ma, K.-C. Chen, S.~Yu, M.~Liu, and R.~Deng, ``Robust decentralized quantum kernel learning for noisy and adversarial environment,'' {\em arXiv preprint arXiv:2504.13782}, 2025.

\bibitem{chen2025consensus}
K.-C. Chen, W.~Ma, and X.~Xu, ``Consensus-based distributed quantum kernel learning for speech recognition,'' in {\em 2025 IEEE International Conference on Acoustics, Speech, and Signal Processing Workshops (ICASSPW)}, pp.~1--5, IEEE, 2025.

\bibitem{chen2025noise}
K.-C. Chen, M.~Prest, F.~Burt, S.~Yu, and K.~K. Leung, ``Noise-aware detectable byzantine agreement for consensus-based distributed quantum computing,'' in {\em 2025 International Conference on Quantum Communications, Networking, and Computing (QCNC)}, pp.~210--215, IEEE, 2025.

\bibitem{chen2025quantum}
K.-C. Chen, S.~Y.-C. Chen, T.-Y. Li, C.-Y. Liu, and K.~K. Leung, ``Quantum machine learning for uav swarm intrusion detection,'' {\em arXiv preprint arXiv:2509.01812}, 2025.

\bibitem{chen2024introduction}
S.~Y.-C. Chen and S.~Yoo, ``Introduction to quantum federated machine learning,'' in {\em Federated Learning}, pp.~311--328, Elsevier, 2024.

\bibitem{chen2021federated}
S.~Y.-C. Chen and S.~Yoo, ``Federated quantum machine learning,'' {\em Entropy}, vol.~23, no.~4, p.~460, 2021.

\bibitem{chehimi2022quantum}
M.~Chehimi and W.~Saad, ``Quantum federated learning with quantum data,'' in {\em ICASSP 2022-2022 IEEE International Conference on Acoustics, Speech and Signal Processing (ICASSP)}, pp.~8617--8621, IEEE, 2022.

\bibitem{mari2020transfer}
A.~Mari, T.~R. Bromley, J.~Izaac, M.~Schuld, and N.~Killoran, ``Transfer learning in hybrid classical-quantum neural networks,'' {\em Quantum}, vol.~4, p.~340, 2020.

\bibitem{liu2024federated}
C.-Y. Liu and S.~Y.-C. Chen, ``Federated quantum-train with batched parameter generation,'' {\em arXiv preprint arXiv:2409.02763}, 2024.

\bibitem{liu2025federated}
C.-Y. Liu, S.~Y.-C. Chen, K.-C. Chen, W.-J. Huang, and Y.-J. Chang, ``Federated quantum-train long short-term memory for gravitational wave signal,'' {\em arXiv preprint arXiv:2503.16049}, 2025.

\bibitem{qu2024quantum}
Z.~Qu, L.~Zhang, and P.~Tiwari, ``Quantum fuzzy federated learning for privacy protection in intelligent information processing,'' {\em IEEE Transactions on Fuzzy Systems}, vol.~33, no.~1, pp.~278--289, 2024.

\bibitem{yun2022slimmable}
W.~J. Yun, J.~P. Kim, S.~Jung, J.~Park, M.~Bennis, and J.~Kim, ``Slimmable quantum federated learning,'' {\em arXiv preprint arXiv:2207.10221}, 2022.

\bibitem{abou2022federated}
Z.~Abou El~Houda, B.~Brik, A.~Ksentini, L.~Khoukhi, and M.~Guizani, ``When federated learning meets game theory: A cooperative framework to secure iiot applications on edge computing,'' {\em IEEE Transactions on Industrial Informatics}, vol.~18, no.~11, pp.~7988--7997, 2022.

\bibitem{ferrag2022edge}
M.~A. Ferrag, O.~Friha, D.~Hamouda, L.~Maglaras, and H.~Janicke, ``Edge-iiotset: A new comprehensive realistic cyber security dataset of iot and iiot applications for centralized and federated learning,'' {\em IEEE Access}, vol.~10, pp.~40281--40306, 2022.

\bibitem{zafari2023resource}
F.~Zafari, P.~Basu, K.~K. Leung, J.~Li, D.~Towsley, and A.~Swami, ``Resource sharing in the edge: A distributed bargaining-theoretic approach,'' {\em IEEE Transactions on Network and Service Management}, vol.~20, no.~4, pp.~4369--4382, 2023.

\bibitem{hejazi2013one}
M.~Hejazi and Y.~P. Singh, ``One-class support vector machines approach to anomaly detection,'' {\em Applied Artificial Intelligence}, vol.~27, no.~5, pp.~351--366, 2013.

\bibitem{xu2023deep}
H.~Xu, G.~Pang, Y.~Wang, and Y.~Wang, ``Deep isolation forest for anomaly detection,'' {\em IEEE Transactions on Knowledge and Data Engineering}, vol.~35, no.~12, pp.~12591--12604, 2023.

\bibitem{lindemann2021survey}
B.~Lindemann, B.~Maschler, N.~Sahlab, and M.~Weyrich, ``A survey on anomaly detection for technical systems using lstm networks,'' {\em Computers in Industry}, vol.~131, p.~103498, 2021.

\bibitem{ergen2019unsupervised}
T.~Ergen and S.~S. Kozat, ``Unsupervised anomaly detection with lstm neural networks,'' {\em IEEE transactions on neural networks and learning systems}, vol.~31, no.~8, pp.~3127--3141, 2019.

\bibitem{malhotra2016lstm}
P.~Malhotra, A.~Ramakrishnan, G.~Anand, L.~Vig, P.~Agarwal, and G.~Shroff, ``Lstm-based encoder-decoder for multi-sensor anomaly detection,'' {\em arXiv preprint arXiv:1607.00148}, 2016.

\bibitem{said2020network}
M.~Said~Elsayed, N.-A. Le-Khac, S.~Dev, and A.~D. Jurcut, ``Network anomaly detection using lstm based autoencoder,'' in {\em Proceedings of the 16th ACM symposium on QoS and security for wireless and mobile networks}, pp.~37--45, 2020.

\bibitem{mezina2021network}
A.~Mezina, R.~Burget, and C.~M. Travieso-Gonz{\'a}lez, ``Network anomaly detection with temporal convolutional network and u-net model,'' {\em IEEE Access}, vol.~9, pp.~143608--143622, 2021.

\bibitem{he2019temporal}
Y.~He and J.~Zhao, ``Temporal convolutional networks for anomaly detection in time series,'' in {\em Journal of Physics: Conference Series}, vol.~1213, p.~042050, IOP Publishing, 2019.

\bibitem{thill2021temporal}
M.~Thill, W.~Konen, H.~Wang, and T.~B{\"a}ck, ``Temporal convolutional autoencoder for unsupervised anomaly detection in time series,'' {\em Applied Soft Computing}, vol.~112, p.~107751, 2021.

\bibitem{blazquez2021review}
A.~Bl{\'a}zquez-Garc{\'\i}a, A.~Conde, U.~Mori, and J.~A. Lozano, ``A review on outlier/anomaly detection in time series data,'' {\em ACM computing surveys (CSUR)}, vol.~54, no.~3, pp.~1--33, 2021.

\bibitem{botterman2022robust}
H.-L. Botterman, J.~Roussel, T.~Morzadec, A.~Jabbari, and N.~Brunel, ``Robust pca for anomaly detection and data imputation in seasonal time series,'' in {\em International Conference on Machine Learning, Optimization, and Data Science}, pp.~281--295, Springer, 2022.

\bibitem{liu2020deep}
Y.~Liu, S.~Garg, J.~Nie, Y.~Zhang, Z.~Xiong, J.~Kang, and M.~S. Hossain, ``Deep anomaly detection for time-series data in industrial iot: A communication-efficient on-device federated learning approach,'' {\em IEEE Internet of Things Journal}, vol.~8, no.~8, pp.~6348--6358, 2020.

\bibitem{deng2021graph}
A.~Deng and B.~Hooi, ``Graph neural network-based anomaly detection in multivariate time series,'' in {\em Proceedings of the AAAI conference on artificial intelligence}, vol.~35, pp.~4027--4035, 2021.

\bibitem{xia2023coupled}
F.~Xia, X.~Chen, S.~Yu, M.~Hou, M.~Liu, and L.~You, ``Coupled attention networks for multivariate time series anomaly detection,'' {\em IEEE Transactions on Emerging Topics in Computing}, vol.~12, no.~1, pp.~240--253, 2023.

\bibitem{polap2023energy}
D.~Po{\l}ap, G.~Srivastava, and A.~Jaszcz, ``Energy consumption prediction model for smart homes via decentralized federated learning with lstm,'' {\em IEEE Transactions on Consumer Electronics}, vol.~70, no.~1, pp.~990--999, 2023.

\bibitem{kaur2024intrusion}
A.~Kaur, ``Intrusion detection approach for industrial internet of things traffic using deep recurrent reinforcement learning assisted federated learning,'' {\em IEEE Transactions on Artificial Intelligence}, 2024.

\bibitem{luo2024securing}
Y.~Luo, X.~Chen, H.~Sun, X.~Li, N.~Ge, W.~Feng, and J.~Lu, ``Securing 5g/6g iot using transformer and personalized federated learning: an access-side distributed malicious traffic detection framework,'' {\em IEEE Open Journal of the Communications Society}, vol.~5, pp.~1325--1339, 2024.

\bibitem{pillutla2022federated}
K.~Pillutla, K.~Malik, A.-R. Mohamed, M.~Rabbat, M.~Sanjabi, and L.~Xiao, ``Federated learning with partial model personalization,'' in {\em International Conference on Machine Learning}, pp.~17716--17758, PMLR, 2022.

\bibitem{tan2022towards}
A.~Z. Tan, H.~Yu, L.~Cui, and Q.~Yang, ``Towards personalized federated learning,'' {\em IEEE transactions on neural networks and learning systems}, vol.~34, no.~12, pp.~9587--9603, 2022.

\bibitem{li2021ditto}
T.~Li, S.~Hu, A.~Beirami, and V.~Smith, ``Ditto: Fair and robust federated learning through personalization,'' in {\em International conference on machine learning}, pp.~6357--6368, PMLR, 2021.

\bibitem{fallah2020personalized}
A.~Fallah, A.~Mokhtari, and A.~Ozdaglar, ``Personalized federated learning: A meta-learning approach,'' {\em arXiv preprint arXiv:2002.07948}, 2020.

\bibitem{yang2023personalized}
L.~Yang, J.~Huang, W.~Lin, and J.~Cao, ``Personalized federated learning on non-iid data via group-based meta-learning,'' {\em ACM Transactions on Knowledge Discovery from Data}, vol.~17, no.~4, pp.~1--20, 2023.

\end{thebibliography}

\end{document}